\documentclass[letterpaper]{article} 
\usepackage{aaai2026}  
\usepackage{times}  
\usepackage{helvet}  
\usepackage{courier}  
\usepackage[hyphens]{url}  
\usepackage{graphicx} 
\usepackage{natbib}  
\usepackage{caption} 
\usepackage{algorithm}
\usepackage{algorithmic}
\usepackage{amsmath}
\usepackage{amssymb}
\usepackage{amsfonts}
\usepackage{tabularx} 
\usepackage{newfloat}
\usepackage{booktabs}
\usepackage{enumitem}
\usepackage{listings}
\usepackage{placeins}
\usepackage{verbatim}

\usepackage{newfloat}
\usepackage{listings}
\DeclareCaptionStyle{ruled}{labelfont=normalfont,labelsep=colon,strut=off} 
\floatstyle{ruled}
\newfloat{listing}{tb}{lst}{}
\floatname{listing}{Listing}
\title{Grounding Vision and Language to 3D Masks \\ for Long-Horizon Box Rearrangement}
\author{
    Ashish Malik, Caleb Lowe, Aayam Shrestha, Stefan Lee, Fuxin Li, Alan Fern
}
\affiliations{
    Electrical Engineering and Computer Science, College of Engineering, Oregon State University\\

    \{malikas, lowecal, aayam.shrestha, leestef, fuxin.li, alan.fern\}@oregonstate.edu
}

\begin{document}

\maketitle

\begin{abstract}
We study long-horizon planning in 3D environments from under-specified natural-language goals using only visual observations, focusing on multi-step 3D box rearrangement tasks. Existing approaches typically rely on symbolic planners with brittle relational grounding of states and goals, or on direct action-sequence generation from 2D vision-language models (VLMs). Both approaches struggle with reasoning over many objects, rich 3D geometry, and implicit semantic constraints. Recent advances in 3D VLMs demonstrate strong grounding of natural-language referents to 3D segmentation masks, suggesting the potential for more general planning capabilities. We extend existing 3D grounding models and propose Reactive Action Mask Planner (RAMP-3D), which formulates long-horizon planning as sequential reactive prediction of paired 3D masks: a "which-object" mask indicating what to pick and a "which-target-region" mask specifying where to place it. The resulting system processes RGB-D observations and natural-language task specifications to reactively generate multi-step pick-and-place actions for 3D box rearrangement. We conduct experiments across 11 task variants in warehouse-style environments with 1-30 boxes and diverse natural-language constraints. RAMP-3D achieves 79.5\% success rate on long-horizon rearrangement tasks and significantly outperforms 2D VLM-based baselines, establishing mask-based reactive policies as a promising alternative to symbolic pipelines for long-horizon planning.
\end{abstract}

\section{Introduction}

Consider a warehouse robot assigned to autonomously rearrange boxes based on natural-language instructions such as \textit{“Move all of the larger boxes on the floor to the back shelves, but don’t stack them more than two high”} or \textit{“group boxes on the pallet by matching color tags, prioritizing yellow.”} To execute these tasks, the robot must interpret underspecified commonsense instructions, understand the 3D geometry and semantics of cluttered scenes, and respect implicit constraints such as keeping boxes compactly and neatly arranged. Even when the low-level system can reliably move a selected box to a specified destination, determining which box to move next and where it should go becomes a complex long-horizon planning problem when the environment contains many objects, relationships, and geometric constraints.


In this work, we focus on the high-level planning problem: given raw visual input and a natural-language command, the planner must output a sequence of abstract pick-and-place actions specifying a source box and a destination region. A common approach is to insert a \emph{symbolic layer} between perception and a symbolic planner, using hand-designed operator vocabularies, scene graphs, or predicate-based representations to generate multi-step plans. However, constructing and maintaining this symbolic layer becomes difficult and unwieldy. The designer must decide which geometric and semantic concepts become explicit predicates, define thresholds and priorities latent in the language, and ensure that these abstractions can be consistently grounded from noisy perceptions. Natural-language goals must also be translated, into formal objectives the planner can interpret. Finally, as tasks vary, these symbolic choices must remain coherent so that new goals and constraints integrate cleanly with existing ones, further increasing the burden of maintaining the symbolic layer.

In contrast, we explore an alternative that avoids engineering an explicit symbolic layer altogether. Our approach is inspired by recent advances in 3D vision-language models (VLMs), which demonstrate strong capabilities in grounding natural-language referents to 3D segmentation masks. Here, we consider whether this capability can extend to direct perception-to-action planning, where the referents are long-horizon task descriptions. Building on these models, we introduce the Reactive Action Mask Planner (RAMP-3D) that treats grounded 3D masks as the action space itself. At each step, the RAMP-3D model processes multi-view RGB-D observations and the natural-language command to produce two masks: a \emph{which-object} mask indicating the next box to pick and a \emph{which-target-region} mask specifying its placement location. These masks are then converted into discrete object and region choices for execution. By repeatedly invoking the planner in this manner, the agent executes the full multi-step plan required for the rearrangement task.

While 3D vision–language models such as UniVLG provide strong single-step grounding of referential expressions, they are not directly suited for long-horizon planning. First, UniVLG’s mask predictions for a referent are independent and in turn insufficient for selecting a coherent pickup box and its corresponding placement region. To address this, we introduce a pair-consistency loss and learned pairing features that encourage the model to produce jointly compatible \emph{which-object} and \emph{which-target-region} masks. Second, while UniVLG is trained on large-scale referent-expression datasets, its performance is very poor on long-horizon task specifications, which are highly out of distribution. We therefore develop a training-data generation methodology that includes: 1) a set of 11 task classes, which cover constraints such as height limits, color priorities, spatial ordering, and accessibility, into diverse instruction sets, 2) action supervision derived from an oracle planner operating on full state information for each task, and 3) LLM-generated natural-language paraphrases that expand the task descriptions in the data set. This multi-view RGB-D scene dataset with 3D action masks paired with varied language enables the extended model to learn meaningful pickup–putdown associations and to interpret the broad range of commonsense directives needed for 3D box rearrangement.



\begin{figure*}[t]
    \centering
    \begin{minipage}[t]{0.22\textwidth}
        \centering
        \includegraphics[width=\linewidth,trim=0 0 0 1cm,clip]{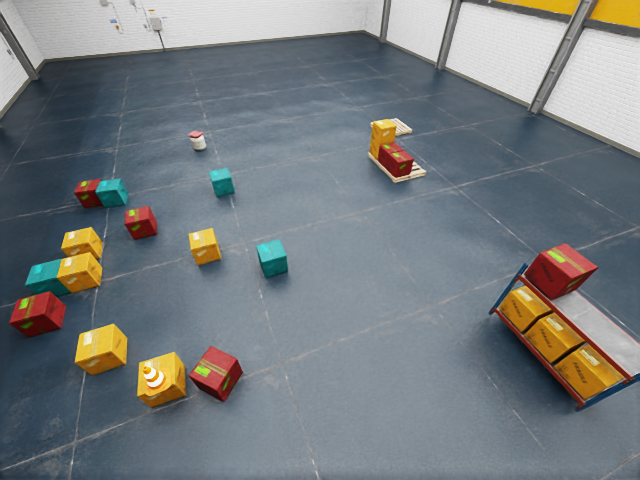}%
        \\[2pt]
        \scriptsize Goal: ``Create stacks of boxes up to a maximum height of 2."
    \end{minipage}\hfill
    \begin{minipage}[t]{0.22\textwidth}
        \centering
        \includegraphics[width=\linewidth,trim=0 0 0 1cm,clip]{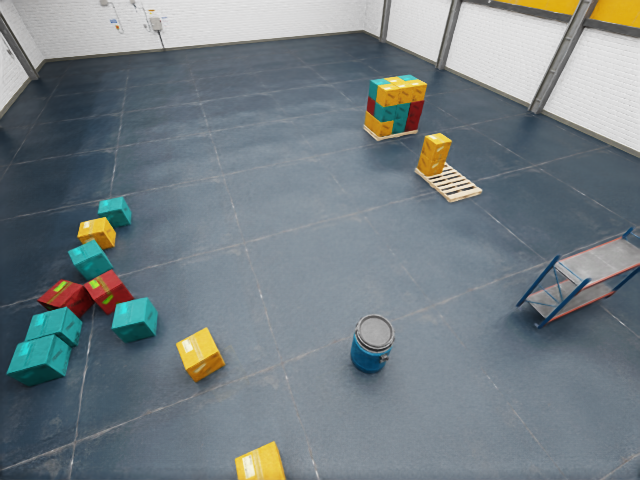}%
        \\[2pt]
        \scriptsize Goal: ``Create stacks of boxes up to height 3, stacking them from right to left."
    \end{minipage}\hfill
    \begin{minipage}[t]{0.22\textwidth}
        \centering
        \includegraphics[width=\linewidth,trim=0 0 0 1cm,clip]{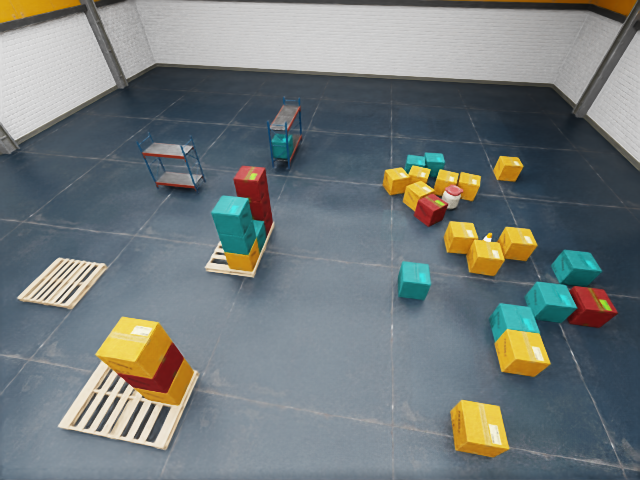}%
        \\[2pt]
        \scriptsize Goal: ``Stack the boxes to a maximum height of 3, completing one stack completely before starting another one."
    \end{minipage}\hfill
    \begin{minipage}[t]{0.22\textwidth}
        \centering
        \includegraphics[width=\linewidth,trim=0 0 0 1cm,clip]{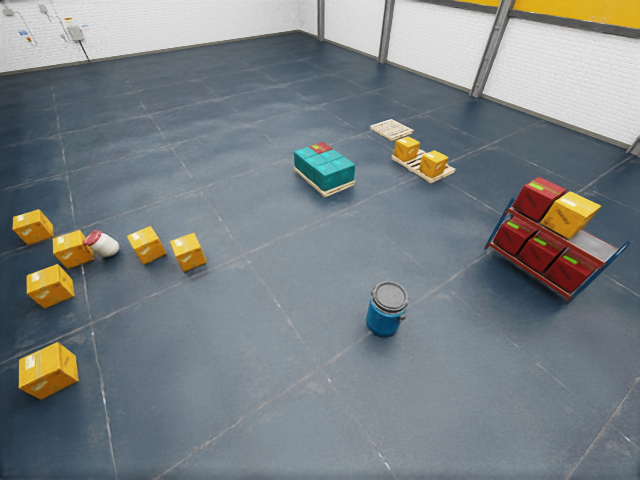}%
        \\[2pt]
        \scriptsize Goal: ``Place all boxes on pallets and shelves without stacking, prioritizing blue boxes first."
    \end{minipage}
    \vskip -0.03in
    \caption{Box-rearrangement in warehouse environment using natural language specified goals.}
    \label{fig:task_variety}
    \vskip -0.15in
\end{figure*}

We evaluate RAMP-3D on visually realistic warehouse-style box-rearrangement tasks involving 1–30 boxes, diverse goal constraints, and long-horizon language specifications. The model shows strong test-set accuracy on predicting individual box–target pairs and achieves up to 90.9\% success when rolled out to produce full plans, with performance varying by scene complexity. Compared to a 2D VLM-based baseline, our 3D mask-based architecture attains substantially higher planning success across all task variants. Overall, these results demonstrate that leveraging 3D grounding VLMs as reactive planners is a promising direction for long-horizon 3D rearrangement, while also revealing failure modes that motivate incorporating additional decision hierarchy or lookahead search. The results further indicate potential for extending this framework to richer 3D manipulation domains. In summary, our key contributions are:

\begin{enumerate}
    \item \textbf{A formulation of 3D rearrangement planning as grounded mask selection.} 
    We introduce a long-horizon planning paradigm that expresses high-level actions as paired 3D ``which-object'' and ``which-region'' masks, avoiding symbolic operators or explicit search.
    \item \textbf{A 3D VLM-based reactive planner with task-specific training.} We develop RAMP-3D by extending a 3D grounding VLM with architectural and training modifications that enable sequential pick-and-place decisions. To our knowledge, this is the first work to investigate whether models designed for 3D referent grounding can be adapted to handle natural-language specifications that require long-horizon planning.
    \item \textbf{A benchmark and empirical evaluation.}
    We provide a diverse suite of language-conditioned 3D box rearrangement tasks and demonstrate strong long-horizon performance compared to a 2D VLM baseline.
\end{enumerate}

\section{Related works}

Our work on long-horizon rearrangement with vision and language inputs spans several families of methods, including symbolic planning interfaces, LLM/VLM-driven decision making, and learned reactive planners. We summarize the most relevant threads below.

\textbf{Symbolic Planning Interfaces.} One strategy is to translate raw vision and language into an explicit symbolic representation that can be consumed by a classical planner. This requires perceiving the scene well enough to populate hand-engineered predicates, scene graphs, or operator vocabularies \citep{jiao2022sequential, li2024llm}. Even assuming perfect semantic reconstruction, representing the geometric constraints of 3D manipulation in formalisms such as PDDL remains difficult, as these languages provide limited support for continuous spatial relations. Task-and-motion planning (TAMP) systems \citep{garrett2021}, such as PDDLStreams \citep{garrett2020}, address some of these limitations by combining symbolic search with numerical reasoning over object and robot poses, but they still depend on a symbolic layer whose expressivity matches the richness of the task.

Even with a sufficiently expressive symbolic layer, translating natural-language goals into these symbolic structures introduces another layer of complexity. The planner must determine which linguistic concepts become explicit predicates, which implicit constraints (e.g., accessibility, ordering, stackability) must be encoded numerically, and how these abstractions should be composed into a coherent goal specification. Recent work explores using LLMs or VLMs to assist with this translation, for example, by inferring symbolic goal states \citep{ding2023}, predicting subgoals \citep{yang2025}, or generating PDDL representations directly from natural language \citep{han2024interpret, huang2024understanding, zhang2024dkprompt}. While promising, these approaches often rely on oracle-level object information and focus on settings where language maps cleanly onto a small set of predicates. As tasks become more varied and constraints more implicit, maintaining a symbolic vocabulary that remains closed under new goals becomes increasingly difficult, especially in 3D rearrangement domains.

\textbf{LLM/VLM-Driven Planning.} Another line of work integrates large language or vision-language models (LLMs/VLMs) directly into the decision loop. SayCan \citep{ahn2022can} is a prototypical example, using an LLM to propose high-level actions while a learned value function grounds them in robot affordances. Later methods treat LLMs/VLMs as critics, re-rankers, or replanners, prompting the model to revise an action sequence when execution fails \citep{mei2024replanvlm}. These approaches have shown promising results in short-horizon tabletop domains but typically rely on carefully engineered prompts tuned to specific classes of tasks. Moreover, the VLMs used in these systems operate on 2D image features, which limits their ability to reason about 3D geometry, occlusion, and volumetric relationships--capabilities that are essential for long-horizon rearrangement.
In contrast, our approach is grounded explicitly in 3D spatial reasoning by leveraging a 3D grounding vision-language model. To the best of our knowledge, this is the first work to use such models for long-horizon planning in visually complex 3D environments.

\textbf{Learning Reactive Policies for Planning.} Learning goal-conditioned reactive policies for classical symbolic planning domains has a long history. Early approaches learned generalized reactive policies by imitating solutions produced by planners \citep{khardon1999,toyer2020} or inducing rules from goals achieved during random walks \citep{fern2004}. These methods replace explicit search at test time but assume a fully symbolic planning model with hand-specified predicates and transition rules. 

More recent work in hierarchical vision–language–action (VLA) models, such as $\pi_0$-style architectures \citep{black2024}, learns end-to-end vision-based policies from demonstrations annotated with high-level abstract actions and corresponding low-level control behaviors. These systems produce abstract actions that are subsequently expanded into motor skills, and have shown strong results in structured tasks. However, they are fundamentally 2D-vision systems that make decisions using only the current egocentric view, without constructing a persistent 3D scene representation. As a result, their ability to reason about occluded objects, geometric structure, and spatially extended environments is limited. Moreover, such models are typically trained on relatively narrow activity domains (e.g., cooking or tool use) with limited natural-language variability, in contrast to our setting where language conveys diverse commonsense constraints for long-horizon 3D rearrangement tasks.

Our approach differs in that we build a full 3D grounding model of the environment and express high-level decisions directly through grounded 3D masks, enabling reactive planning informed by the accumulated 3D structure of the scene rather than a 2D image views.

\section{Problem Formulation}

We consider the task of long-horizon box rearrangement from 3D visual input and natural-language goals. To isolate the core challenges, we focus on a warehouse-style \emph{box rearrangement} domain. This setting contains rich 3D structure, multi-step spatial dependencies, and a broad range of linguistically underspecified goals, yet limits the semantic breadth of object types. Although more general semantic domains are possible, box rearrangement already captures many of the complexities of long-horizon 3D manipulation including spatial relations, accessibility constraints, ordering requirements, and sequential dependencies.

At each high-level planning step, we assume that the planner receives a set of posed RGB-D images that collectively cover the workspace. These images may come from mobile robots carrying calibrated cameras,  static overhead cameras, or  a combination of sources. After each executed pick-and-place action, the visual observations are refreshed to reflect the updated scene. Each camera view is associated with known intrinsics and an estimated camera-to-world pose, obtained either from robot SLAM or from fixed calibration, enabling consistent 3D reconstruction across views.

Formally, each problem instance consists of a set of boxes $\mathcal{B}$, pallets $\mathcal{P}$, and shelves $\mathcal{S}$ arranged on a planar floor. At time step $t$, the environment occupies an unobserved state $s_t$. The planner receives a natural-language goal $g$ together with a collection of $V$ RGB-D observations:
\[
O_t = \Big( \{(I_t^v, D_t^v, K^v, T^v_{c\rightarrow w})\}_{v=1}^V,\; g \Big),
\]
where $I_t^v \in \mathbb{R}^{H \times W \times 3}$ is the RGB image for view $v$,  
$D_t^v \in \mathbb{R}^{H \times W}$ is the corresponding depth map,  
$K^v$ is the intrinsics matrix,  
and $T^v_{c\rightarrow w} \in \mathrm{SE}(3)$ is the camera-to-world pose.

Given these observations, the planner must output an abstract pick-and-place action
\[
a_t = (b_t, r_t),
\]
where $b_t \in \mathcal{B}$ is the selected box to move and $r_t$ is the target region for placement (e.g., a pallet region, shelf region, another box, or a floor location). Executing $a_t$ induces a transition $s_t \rightarrow s_{t+1}$, after which the planner receives updated observations $O_{t+1}$ and selects the next action. This process continues until the planner determines that the task goal has been satisfied. Our objective is to design a planner that performs this perception-to-action mapping directly from 3D observations and the natural-language command. Importantly, the planner is not provided with an explicit symbolic representation of the scene or any semantic image annotations—only raw visual input.

\begin{figure*}[!ht]
    \centering
    \includegraphics[width=\textwidth]{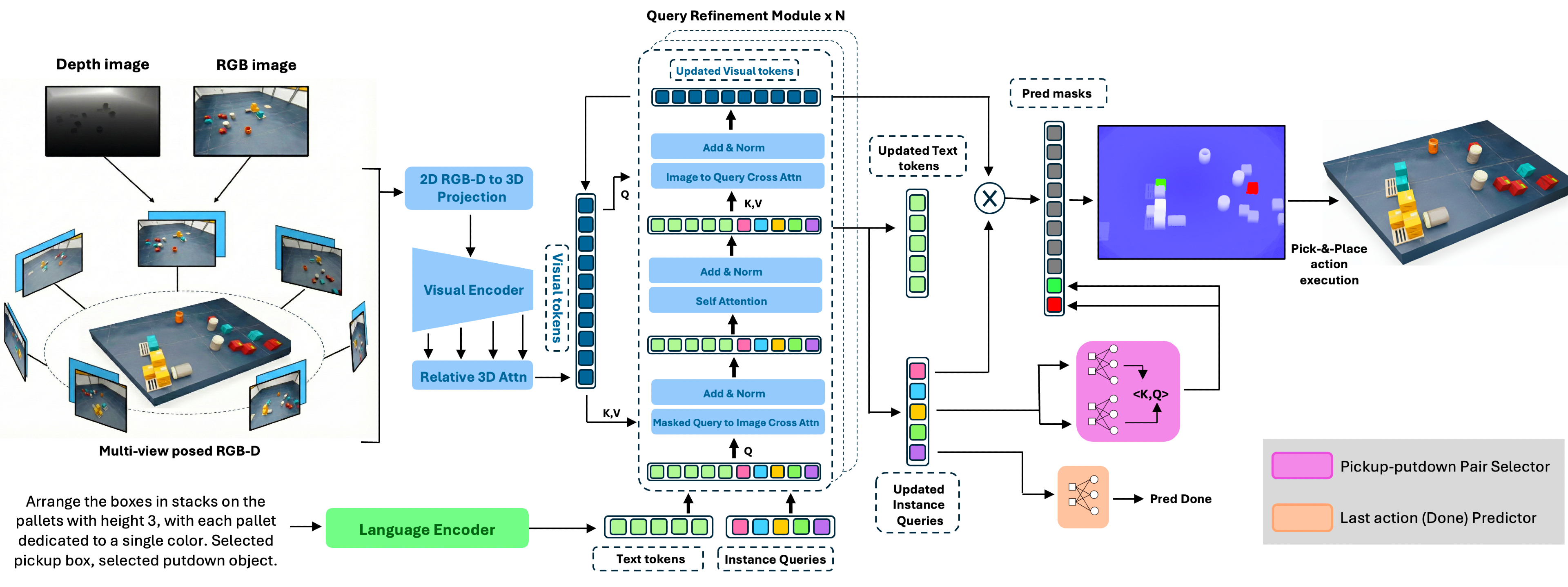}
    \vskip -0.03in
    \caption{Overview of RAMP-3D,  built on UniVLG. The model takes posed multi-view RGB-D observations and a natural-language goal as inputs, encodes visual features with a transformer backbone and voxel-based 3D fusion, and encodes the goal with a frozen text encoder. A transformer decoder with learnable queries attends jointly to visual and text  tokens and are augmented with pair-contrastive pickup–putdown embeddings and a binary “done’’ head. At each planning step, the model outputs a pair of 3D masks indicating the pickup target and target region and a termination probability. The predicted masks are projected back to instance IDs to yield box–region actions. The planner is run iteratively for long-horizon box rearrangement.}
    \vskip -0.15in
    \label{fig:mainFig}
\end{figure*}

\section{Reactive Action Mask Planner (RAMP-3D)}


Figure~\ref{fig:mainFig} illustrates our RAMP-3D model, which builds on the pre-trained UniVLG architecture~\citep{jain2025unifying}, a state-of-the-art model for grounding natural-language referent expressions in 3D scenes. RAMP-3D adapts this grounding capability to produce sequential pick-and-place actions for long-horizon box rearrangement. We first provide a brief overview of the UniVLG architecture, referring the reader to the original work for full technical details. We then describe the extensions required to support action-mask prediction, followed by our training methodology in Section~\ref{sec:training}.

\subsection{UniVLG Overview}

UniVLG takes as input a natural-language query and $N$ posed RGB-D views of a scene, and predicts language-conditioned segmentation masks over 3D points corresponding to the referenced objects. It begins by converting the multi-view images into a unified set of 3D visual tokens. Each RGB-D view is first passed through a transformer-based image backbone that produces multi-resolution feature maps. Using the depth values and known camera poses, every image pixel is then back-projected into 3D space, yielding a point cloud with associated visual features. 

To improve computational efficiency, these 3D points can optionally be voxelized 
so that features are aggregated within each voxel. A multi-view deformable attention module further processes these point/voxel tokens so that features of each token encode both local appearance and  3D geometric relationships across views, resulting in a collection of geometry-aware 3D visual tokens, each associated with a 3D positional encoding.

Language instructions are encoded using a frozen text encoder into contextual token embeddings, with the tokens corresponding to referring phrases identified for supervision. A transformer-based mask decoder maintains $Q$ learnable queries, which iteratively update through self-attention across queries and cross-attention to both the 3D visual tokens and the language embeddings. After several decoding layers, for each query UniVLG outputs:
\begin{itemize}
    \item \textbf{Text span logits:} computed by comparing the updated query embedding to language embeddings, indicating which tokens (referenced object) in the input sentence the query refers to.
    \item \textbf{Mask logits:} computed by projecting the query embedding onto the 3D visual tokens, producing a per-point mask over the scene for the 
    referenced object.

\end{itemize}
For our mask-based reactive planner, we employ Swin Transformer\citep{liu2021swin} as the visual backbone and RoBERTa-base \citep{liu2019roberta} as the text encoder.

\subsection{Extending UniVLG to RAMP-3D}

We introduce two architectural modifications to the UniVLG decoder that adapt the model to the box-rearrangement planning setting. These modifications allow the model to produce \emph{paired} pickup-putdown predictions and to determine when the high-level plan should terminate.

\paragraph{Pair-contrastive features.} A key requirement of our planner is that it must output a \emph{pair} of consistent predictions: a pickup target and a corresponding putdown region that together define a valid action. To encourage the model to learn such structured dependencies, we introduce a pair-contrastive module built on top of the decoder queries.

Concretely, we introduce two sets of learnable linear projections, $\mathbf{W}_q^{\text{pair}}$ and $\mathbf{W}_k^{\text{pair}}$. For each decoder query $\mathbf{z}_i$, we compute:
\[
\mathbf{q}_i = \mathbf{W}_q^{\text{pair}}\mathbf{z}_i, \qquad 
\mathbf{k}_i = \mathbf{W}_k^{\text{pair}}\mathbf{z}_i,
\]
where $\mathbf{q}_i$ serves as a \emph{pickup embedding} and $\mathbf{k}_i$ serves as a \emph{putdown embedding}. For query indices $\mathcal{J}$, we compute pairwise compatibility scores:
\[
s_{ij} = \langle \mathbf{q}_i, \mathbf{k}_j \rangle, \qquad (i,j)\in\mathcal{J},\; i\neq j.
\]

At inference time, these scores guide the joint selection of pickup and putdown queries. We first identify a small set of top-$K$ (with $K=5$) high-confidence pickup candidates, and for each candidate compute compatibility against all putdown queries. The final action is selected by maximizing combined classification confidence and pairwise compatibility.

\paragraph{Last-action predictor.}
We add a lightweight ``done'' head enabling the model to signal when no further moves are required. This head is implemented by a single-hidden-layer MLP with 256 hidden units that takes all decoder queries as input and outputs a scalar termination probability.

\paragraph{Mask-space action representation.} Our planner operates directly in \emph{mask space}. Let $\mathcal{X}$ denote the discrete 3D point cloud constructed from the multi-view RGB-D observations. At each planning step, the model predicts two 3D masks:
\[
M_t^{\text{pick}},\quad M_t^{\text{target}} \in \{0,1\}^{\mathcal{X}}
\]
corresponding to the pickup object and its putdown region. These masks are then translated into high-level action commands for a pick-and-place controller. In our simulation experiments, this corresponds to projecting the masks back to object IDs (the object with highest mask coverage) and geometric target regions, yielding the discrete action $(b_t, r_t)$ for execution.  In a robotics deployment, the masks would be translated to robot relative coordinates and passed to a pick-and-place controller that is trained or program to interpret the provided coordinates. 

This mask-based formulation turns the 3D grounding model into a semantics-to-geometry planning interface: predicted masks inherently encode geometric feasibility and occupancy constraints, allowing the planner to reason over 3D structure without an explicit symbolic abstraction.

\subsection{Training Data Generation}
\label{sec:training}


We generate training data using the IsaacSim simulator, collecting multi-view RGB-D observations, instance annotations, and ground-truth pick-and-place actions produced by an oracle planner for box rearrangement tasks.


\paragraph{Simulation Environment: } Scenes are defined over a 12 m × 12 m warehouse with a planar floor and populated with:

\begin{itemize}[leftmargin=0.9em]
\item \textit{Pallets}: serve as single layered placement surface for boxes. Each scene contain up to 3 pallets which comes in one of two sizes: small (accommodating roughly 4 boxes per level in a $2 \times 2$ arrangement) and large (accommodating roughly 6 boxes in a $2 \times 3$ arrangement). Boxes can be stacked up to three levels high on each pallet.
\item \textit{Shelves}: provide multi-layer storage. Scenes contain up to 2 shelves, either a small shelf (2 boxes per layer), or a large shelf (3 boxes per layer). Each shelf has 2 horizontal layers and boxes are not stacked within a layer.
\item \textit{Boxes}: are the objects to be rearranged which share identical physical dimensions and come in 3 colors (red, blue, yellow). Each Scene contains up to 30 boxes total.
\item \textit{Distractors} add visual clutter without being manipulated. Scenes include up to 4 distractors sampled from 6 types: five distinct barrel types and a traffic cone.
\end{itemize}

\paragraph{Scene Initialization:} Pallets, shelves, and boxes spawn in separate designated floor regions, with distractors spawning alongside boxes. We ensure collision-free placement of pallets and shelves, applying random yaw rotations in \{0°, 90°\} to increase visual diversity. Boxes and distractors are spawned without collision awareness and settle to arbitrary orientations through physics simulation before data collection begins. Multi-view observations are captured using V = 30 cameras positioned on a circular trajectory (radius 5.0 m, height 5.0 m) pointing toward the scene center. We record RGB images, depth maps, and instance segmentations—the latter used to generate ground-truth masks for supervision, though not provided as input to the model.

\subsubsection{Task Variants:}
We define 11 task variants that capture diverse commonsense constraints in warehouse rearrangement. These tasks constrain both final configurations and intermediate placement ordering. Box stacking is permitted up to three boxes high unless specified otherwise. We outline these tasks in Table \ref{tab:task_variants}. For each task type we create three natural-language goal templates with randomly sampled parameters (e.g., maximum stack height, prioritized box type, preferred pallet size) that capture the task constraints. 

\begin{table}[t]
\centering
\scriptsize
\setlength{\tabcolsep}{3em} 
\begin{tabularx}{\columnwidth}{@{}l@{\hspace{0.85em}}X@{}} 
\toprule
\textbf{Task Variant} & \textbf{Description} \\
\midrule
Basic placement          & Standard box emplacement with stack height limits. \\
Box type prioritization  & Prioritize designated box types before others. \\
Shelf prioritization     & Fill all shelves completely before placing boxes on pallets. \\
Pallet prioritization    & Fill pallets before placing boxes on shelves. \\
Placement ordering       & Follow prescribed left-to-right or right-to-left ordering. \\
Size-based prioritization& Fill pallets or shelves of specified size before others. \\
Avoid stacking           & Stack boxes only when no unstacked locations remain. \\
Homogeneous stacks       & Each stack must contain a single type of boxes. \\
Box-type segregation     & Each pallet or shelf must contain a single type of boxes. \\
Finish stack first       & Finish current stack before selecting other locations. \\
Box accessibility        & Keep selected box types accessible by placing them on top of stacks
                           or in homogeneous stacks. \\
\bottomrule
\end{tabularx}
\caption{
\footnotesize
List of task variants. Each variant captures
different commonsense constraints on final configurations and intermediate
placement ordering. All tasks permit stacking up to three boxes high where allowed. Refer to Fig \ref{fig:task_variety} for some examples.}
\label{tab:task_variants}
\vskip -0.15in
\end{table}

\subsubsection{Data Collection Procedure:} Training data are generated by constructing and solving box-rearrangement tasks in IsaacSim. For each task instance, we sample a task variant, instantiate a  corresponding scene, and apply a task-specific oracle planner to compute a sequence of ground-truth pick-and-place actions. Note that for most steps there exist multiple valid actions and our oracle selects a single ground truth action with the smallest pickup-to-putdown distance.

After each oracle action, we record multi-view RGB images $I_t^v$, depth maps $D_t^v$, instance segmentations $Z_t^v$ (used only for deriving supervision), and camera poses $T^v_{c\rightarrow w} \in \mathrm{SE}(4)$, along with camera intrinsics $K^v$ and the natural-language goal~$g$. 
Each such snapshot forms a  training example for predicting the next abstract action masks, where ground-truth masks are derived from the instance segmentations and the oracle-selected pick-and-place action. Our resulting dataset comprises approximately 9.5k unique scene samples of which we hold out 2.2K for testing data.

\subsubsection{Natural Language Data Augmentation:} To improve language generalization, we apply \emph{language augmentation} using LLM-based paraphrasing. Each generated training scenario is originally associated with a template-based natural-language goal. We augment the training data beyond the templates by using the gpt-oss-120b model \citep{agarwal2025gpt} to generate up to three semantically equivalent variants for each training instance. The model is prompted with the original natural-language goal and object descriptions, requesting paraphrases that preserve the intended meaning while varying linguistic form. Since UniVLG requires explicit referring phrases for supervision, we additionally append the phrase ``selected pickup box, selected putdown object’’ to each augmented instruction to provide grounding signals for both pickup and putdown mask prediction.

\subsubsection{Auxiliary Prediction Tasks:} Beyond basic pick-and-place supervision, we also generate other language-conditioned tasks at each planning step to encourage the model to learn more fine-grained scene structure. These auxiliary tasks (sampled at a 10\% rate during training) include identifying: (1) all free placement cells, (2) all placed boxes, (3) boxes in stacks (stack height $>1$), (4) accessible boxes, and (5) all unplaced boxes.

\section{EXPERIMENTAL RESULTS}


\subsubsection{Baseline: 2D-pointer}

\begin{figure*}[t]
    \centering
    \small
    \setlength{\tabcolsep}{6pt}

    \begin{minipage}[b]{0.48\textwidth}
        \centering
        \begin{tabular}{ccc}
            \toprule
            Num boxes & RAMP-3D (\%) $\uparrow$ & 2D-pointer (\%) $\uparrow$ \\
            \midrule
            1--10    & 97.9 & 42.4 \\
            11--20   & 97.1 & 39.2 \\
            21--30   & 94.4 & 43.1 \\
            \midrule
            Aggregate  & 96.8 & 41.4 \\
            \bottomrule
        \end{tabular}
        \captionof{table}{One-step plan validity in \emph{snap-to-target} mode.}
        \label{tab:onestep}
    \end{minipage}
    \hfill
    \begin{minipage}[b]{0.48\textwidth}
        \centering
        \begin{tabular}{ccc}
            \toprule
            Putdown target & RAMP-3D (m) $\downarrow$ & 2D-pointer (m) $\downarrow$\\
            \midrule
            Pallet cell & 0.167 $\pm$ 0.11  & 0.125 $\pm$ 0.06  \\
            Box            & 0.124 $\pm$ 0.08  & 0.132 $\pm$ 0.07  \\
            \midrule
            Aggregate      & 0.142 $\pm$ 0.09 & 0.129 $\pm$ 0.07 \\
            \bottomrule 
            \vspace{-0.6em}
        \end{tabular}
        \captionof{table}{Putdown target placement error in \emph{free-form} mode.}
        \label{tab:putdown_error}
    \end{minipage}

    \label{fig:onestep_putdown}
    \vskip -0.1in
\end{figure*}


Designing a meaningful baseline for long-horizon box rearrangement is challenging, as many intuitive approaches either require substantial engineering effort or fail at necessary 3D spatial reasoning. We evaluated several plausible approaches before settling on our final baseline. First, we attempted a pure LLM-style planner inspired by SayCan~\citep{ahn2022can}, prompting an LLM to output pickup-putdown pairs directly, given privileged, ground-truth scene information. However, the model struggled to consistently satisfy accessibility and stacking constraints. We next explored whether a state-of-the-art 2D vision–language pointing model (Molmo~\cite{deitke2024molmo}) could identify the next object and placement region, but such models are designed for referential grounding rather than task-driven spatial reasoning and performed poorly.


These observations motivated a more structured 2D composition, leading to our final baseline, 2D-pointer. In stage one, GPT-5.1 serves as a high-level planner, producing concise relational descriptions of the box and target region based on language goal, 5 RGB images (4 orthogonal views plus  top-down), and aided by 2 in-context examples. In stage two, Molmo-72B grounds these descriptions in the top-down image, ``pointing’’ to the corresponding locations. We then recover discrete box and region by mapping the 2D points to instance IDs via the instance map. This baseline probes how far one can push a describe–then–ground pipeline using strong, off-the-shelf 2D VLMs. We evaluate it only in settings without shelves, since the top-down pointing mechanism is not well suited to multi-layer putdown surfaces.

\subsubsection{Action Execution Modes:} Before evaluating our planner models, we specify the assumptions on the  low-level pickup-putdown controller. We assume the low-level controller can execute the putdown so as to align the box orientation with that of the target region, leading to a neatly packed arrangement. We consider two execution modes: (i) \emph{snap-to-target} mode, in which the controller is provided with the ideal placement location on the predicted putdown target, and (ii) \emph{free-form} mode, in which the controller executes the putdown exactly at the location predicted by the planner. 

For RAMP-3D, the putdown location in \emph{free-form} mode is calculated by averaging the 3D coordinates of all points in the predicted putdown target mask. For the 2D-pointer baseline, the putdown location is computed by backprojecting the pointed 2D location in the top-down view into 3D using the corresponding depth image, camera intrinsics, and camera pose. A known limitation of the UniVLG model is that output masks are not always contiguous components with small noisy disconnected segments. We use standard post processing of each predicted mask using the DBSCAN clustering algorithm to retain only the largest cluster of points.

\

 


\subsubsection{One-Step Prediction Performance:} We first evaluate RAMP-3D and 2D-pointer on their single-step action accuracy. All accuracies are measured with respect to the 2.2K held-out test instances, noting that 2D-pointer is unable to be evaluated on instances involving shelves. A pick-and-place action is considered to be valid if it satisfies all task constraints, including both the final configuration requirements and any object and placement ordering constraints. 

Table \ref{tab:onestep} reports the \emph{snap-to-target} mode's percentage of valid one-step actions over all the test instances and aggregated by number of boxes in a scene. We see that RAMP-3D achieves $96.8\%$ versus 2D-pointer's $41.4\%$. In analyzing the outputs of 2D-pointer, we found that its most frequent failure mode was pointing to unplaced boxes for both the pickup and putdown locations. We believe this arises because pallet placement cells lack distinct, object-like visual cues, whereas MOLMO is primarily trained to point to visually distinctive objects. As a result, the model tends to default to salient box regions rather than the intended placement locations. Finally, we notice that RAMP-3D shows only a modest decrease in performance as scenes become more cluttered with number of boxes, maintaining relatively consistent accuracy across the full range of problem sizes. 




We now evaluate the \emph{free-form} action-selection mode to compare the raw putdown target localization accuracy of the two models. Here we only consider actions with target boxes or pallets, since 2D-pointer is not evaluated on shelf targets. Because localization error is only meaningful for \emph{valid} actions, all reported numbers are averaged exclusively over valid predictions, meaning that RAMP-3D is evaluated on a substantially larger set of examples than 2D-pointer. Table~\ref{tab:putdown_error} reports the Euclidean distance between each model’s predicted putdown location and the ground-truth target center. Overall the localization accuracy of both systems is quite small and within a reasonable range for a heuristic \emph{snap-to-target} pick-and-place controller. 


\subsubsection{Mask Quality Evaluation:} 
\begin{figure}[t]  
    \centering
    \begin{minipage}[b]{0.26\textwidth}
        \centering
        \includegraphics[width=\linewidth]{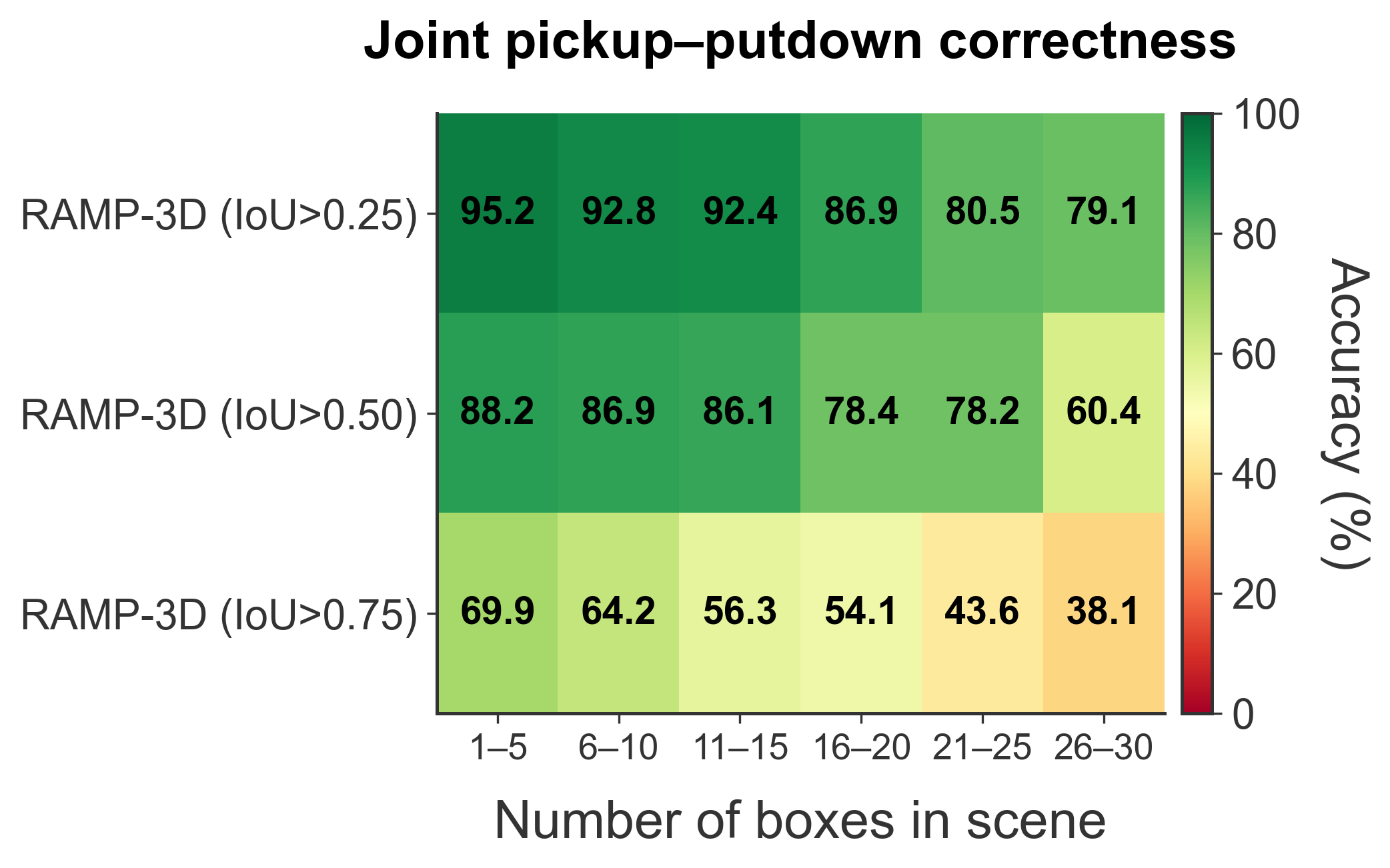}
    \end{minipage}
    \hfill
    \begin{minipage}[b]{0.20\textwidth}
        \centering
        \includegraphics[width=\linewidth]{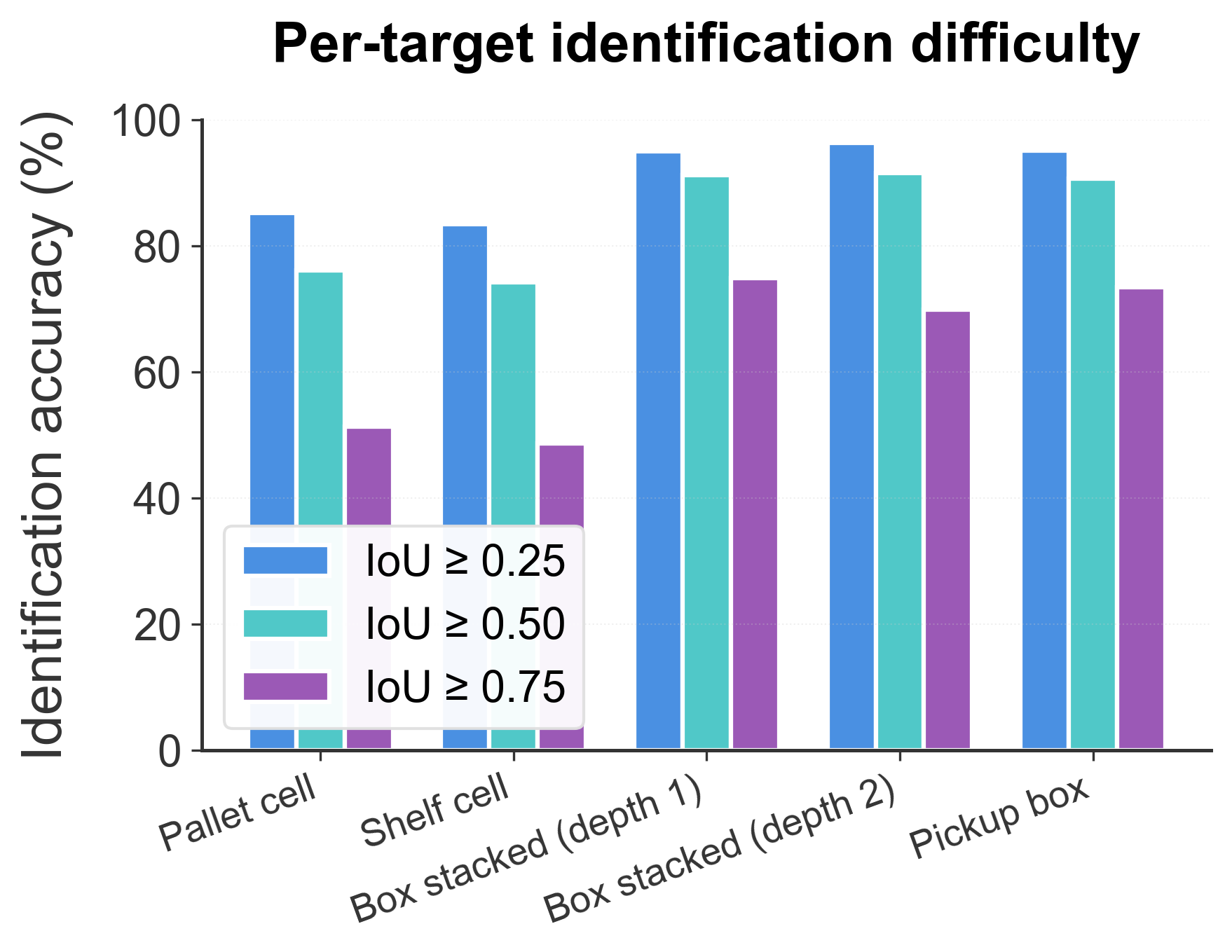}
    \end{minipage}

    \caption{Joint target identification accuracy by number of boxes for RAMP-3D \& 2D-pointer and target identification difficulty for RAMP-3D broken down by the type of targets.}
    \label{fig:targets}
    \vskip -0.15in
\end{figure}

We now evaluate how accurately the masks produced by RAMP-3D correspond to ground-truth 3D masks, providing a direct measure of how well the model captures the semantic structure of the environment. We compute the intersection-over-union (IoU) between the predicted and ground-truth masks for both the pickup box and the putdown region. A prediction is considered correct at IoU threshold $\tau$ only if both masks achieve IoU $\ge \tau$. Figure~\ref{fig:targets} reports these joint target-identification accuracies across scenes grouped by the number of boxes.


In computer vision benchmarks, $\tau=0.25$ is often considered a strong IoU threshold, and we observe that RAMP-3D maintains high joint accuracy at this level across all scene sizes. As expected, accuracy decreases at more stringent thresholds, but remains above 50\% for $\tau = 0.5$ and $\tau = 0.75$ except in the most cluttered scenes. To understand the specific sources of error, we also report per-target accuracies (Fig.~\ref{fig:targets}). The most challenging predictions consistently involve placement cells. We hypothesize that this gap arises because individual placement cells lack distinctive, instance-specific visual cues in RGB-D space, making them inherently harder to localize than boxes.

We notice that RAMP-3D maintains high joint target-identification accuracy, especially in sparsely cluttered scenes. On the other hand, even when counting any feasible pickup-putdown pair detection as success, the 2D pointing pipeline produces a viable action on average in only $33.9\%$ of test samples and exactly matches the oracle pickup and putdown in just $2.0\%$ of cases. In contrast, RAMP-3D correctly identifies both targets with IoU~$\ge 0.25$ in $~89\%$ of scenes on average. While the metrics are not directly comparable, this analysis suggests that operating directly in 3D mask space makes it substantially better to consistently select correct pickup-putdown pairs.

\begin{figure*}[!ht]
    \centering
    \includegraphics[width=0.92\textwidth]{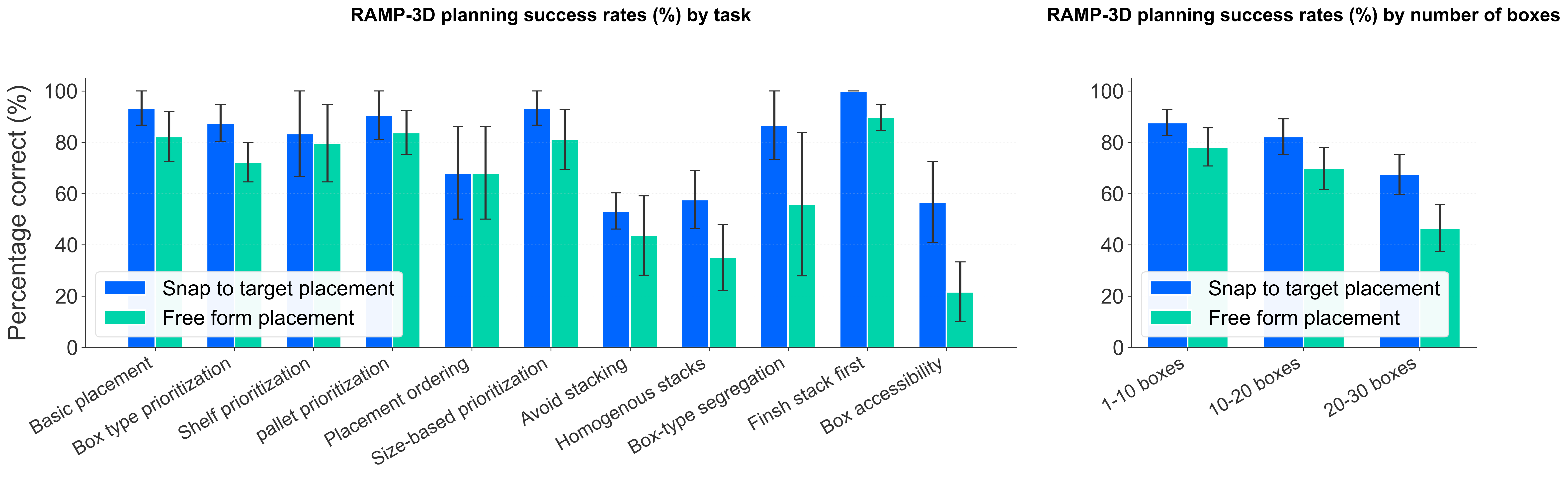}
    \vspace{-1em}
    \caption{Long-horizon plan success rates for RAMP-3D under \emph{snap-to-target}  and \emph{free-form} action execution. Bars show the percentage of rollouts without any invalid actions, broken down by number of boxes per scene and task variants. RAMP-3D attains 79.5\% success on average in \emph{snap-to-target}  mode and 66.5\% in \emph{free-form} mode.}
    \label{fig:planPerformance}
    \vskip -0.15in
\end{figure*}

To understand which targets are most challenging, we also report per-target identification accuracy in Fig.~\ref{fig:targets}. The lowest accuracies consistently correspond to placement cells. We hypothesize that this performance gap between the cells and boxes arises because individual placement cells do not exhibit strong, instance-specific visual cues in RGB-D space.

Taken together, these results indicate that RAMP-3D not only predicts valid actions at a much higher rate than the 2D-pointing baseline, but also produces high-IoU masks that reflect strong 3D grounding and scene understanding.

\subsubsection{Multi-Step Plan Evaluation}

We now evaluate RAMP-3D as a full long-horizon planner by rolling it out to completion on held-out scenes. Note that due to 2D-pointers high single-step error rate it is unable to successfully produce full plans and is not included in the multi-step evaluation. 

Fig \ref{fig:planPerformance} breakdown the planning performance across 200 randomly generated test scenarios, spanning all 11 task variants and scenes with 1–30 boxes. A plan rollout is considered successful when it terminates in a configuration that satisfies all task constraints (including ordering and intermediate constraints). Overall, RAMP-3D achieves $79.5\%$ plan success in the \emph{snap-to-target} mode and $66.5\%$ success in the \emph{free-form} mode. There is a significant gap between \emph{snap-to-target} and \emph{free-form}. In \emph{free-form} mode, predicted masks are converted to continuous 3D putdown locations, which can produce slightly shifted, yet still geometrically valid, placements compared to the discretized placement cells  used in  training. These small  deviations accumulate over multiple steps, pushing scenes out of the training distribution and amplifying failures.


Plan success degrades systematically with increasing scene complexity. With 1–10 boxes, RAMP-3D succeeds on $87.8\%$ of rollouts with \emph{snap-to-target} and $81.1\%$ in \emph{free-form} mode, whereas for the most cluttered 20–30 box scenes, success decreases to $66.7\%$ and $45.6\%$, respectively. This suggests that while the RAMP-3D remains robust in moderately cluttered scenes, the combination of longer horizons and more intricate object interactions makes it increasingly likely to violate ordering or stacking constraints.  

Performance also varies across task families. Tasks with relatively local structure, such as \emph{Basic placement}, \emph{Size-based prioritization}, and \emph{Finish stack first}, attain high success rates, indicating that RAMP-3D can reliably realize goals that primarily require filling particular pallets or shelves in a prescribed order. In contrast, tasks that encode more global or fragile constraints, \emph{Avoid stacking}, \emph{Homogeneous stacks}, and especially \emph{Box accessibility} are substantially harder. These require anticipating long-term consequences of actions, which is challenging for a strictly reactive policy with no lookahead. Finally, we see the gap between \emph{snap-to-target}  and \emph{free-form} execution becomes larger for the most difficult tasks due to the longer horizons which can result in highly out-of-distribution scenes. 


\subsection{Language Generalization:} All language utterances used for training are produced by the paraphraser using the three hand-crafted natural-language templates per task type. This makes it difficult to guarantee that a specific linguistic variation used in testing are never encountered during training. Here, we hold out three canonical templates during training and use them as a test set to probe the robustness of the model to new language realizations of the same underlying goal. The percentage points change of valid one-step actions for \emph{snap-to-target} execution with these templates as task instructions are listed in Table \ref{tab:ablate}. The results shows a minor decrease in performance relative to the instructions from the paraphraser. However, the overall degradation is modest, indicating strong generalization. Larger scale training on wider varieties of paraphrasers will likely improve this dimension of performance. 

\subsubsection{Auxiliary Task Ablation:} We also evaluate the effect of training with and without the auxiliary tasks.  Table \ref{tab:ablate} shows that  removing the auxiliary objectives from the training data yields a $1.7 \%$ decrease in the one-step action validity. This suggests that auxiliary tasks provides additional albeit relatively modest learning signal.

\begin{table}[t]
    \centering
    \small
    \setlength{\tabcolsep}{6pt}
    \begin{tabular}{lcccc}
    \toprule
    \textbf{Experiment} & \textbf{1-10} & \textbf{11-20} & \textbf{21-30} & \textbf{Agg} \\
    \midrule
    Lang. generalization gap & -4.3 & -5.6 & -2.3 & -4.4 \\
     (held-out templates vs. main) & & & & \\
    \addlinespace
    Auxiliary objective gap & -2.4 & -2.1 & -0.8 & -1.9 \\
    (w/o auxiliary vs. main) & & & & \\
    \bottomrule
    \end{tabular}
    \caption{
    Language generalization and auxiliary task ablation. Performance gaps (in percentage points) are relative to the main RAMP-3D model   . 
    }
    \label{tab:ablate}
    \vspace{-1.5em}
\end{table}




\section{Conclusion and future works}

Our work demonstrates that 3D grounding models provide a viable alternative to current 2D pipelines for long-horizon spatial rearrangement. By formulating actions as paired 3D pickup–putdown masks, we show that a learned 3D grounding model can achieve high one-step validity and strong multi-step success across a diverse language-conditioned box rearrangement tasks, without explicit symbolic operators or search. 

At the same time, several limitations highlight opportunities for improvements. First, while our scenes contain a relatively large number of objects, their visual and geometric complexities are fairly limited. We hypothesize that scaling the dataset variety and size along with model-size will provide significantly improved performance and generalization capabilities. Second, the planner is strictly reactive without multi-step lookahead. This works well for many commonsense rearrangement goals where local improvements are aligned with global objectives, but is likely insufficient for non-commonsense or tightly constrained tasks. Another direction is to couple mask-based 3D grounding with optimization tools such as bin-packing, combinatorial solvers or lookahead search.

Finally, we assumed that any valid pickup–putdown pair selected by the high-level planner are executable. In realistic robotic systems, reachability, collision avoidance, and sensing errors will restrict which actions are actually feasible. Incorporating robot-specific feasibility into the training signal, or embedding RAMP-3D within a closed-loop control stack that can detect infeasible actions and recover from failed executions is an important direction for future work.

\section{Acknowledgement}
This work is supported by NSF Award 2321851, DARPA contract HR0011-24-9-0423, and the NVIDIA Academic Grant Program.

\bibliography{aaai2026}
\newpage

\onecolumn
\section*{Appendix}
\subsection{Scene evolution examples}

\begin{figure}[H]
    \centering
    \begin{minipage}{0.25\textwidth}
        \centering
        \includegraphics[width=\linewidth,trim=0 0 0 1cm,clip]{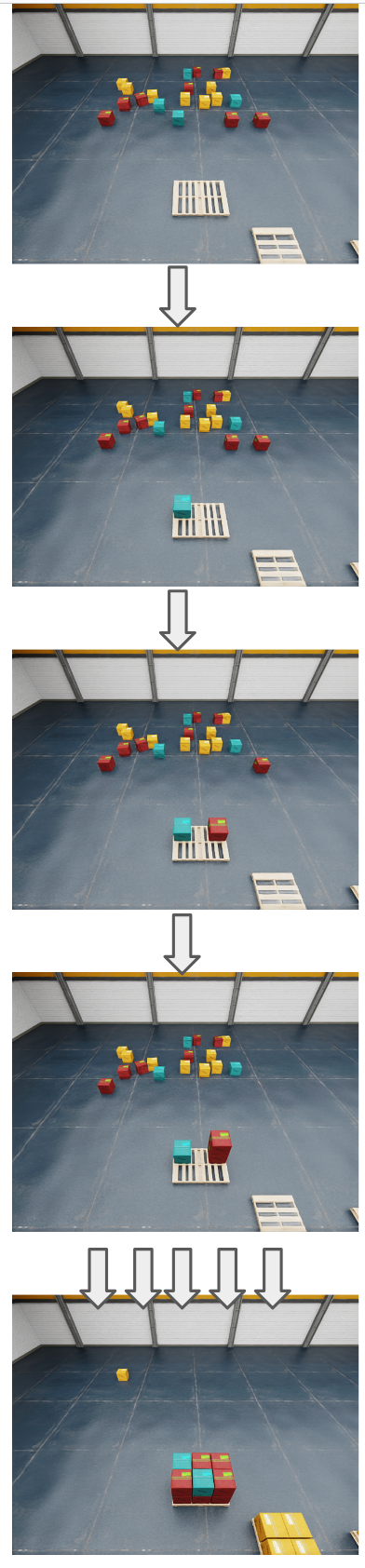}%
        \\[2pt]
        \scriptsize Goal: ``Create stacks of boxes up to height 2, with yellow boxes easily reachable."
    \end{minipage}\hfill
    \begin{minipage}{0.25\textwidth}
        \centering
        \includegraphics[width=\linewidth,trim=0 0 0 1cm,clip]{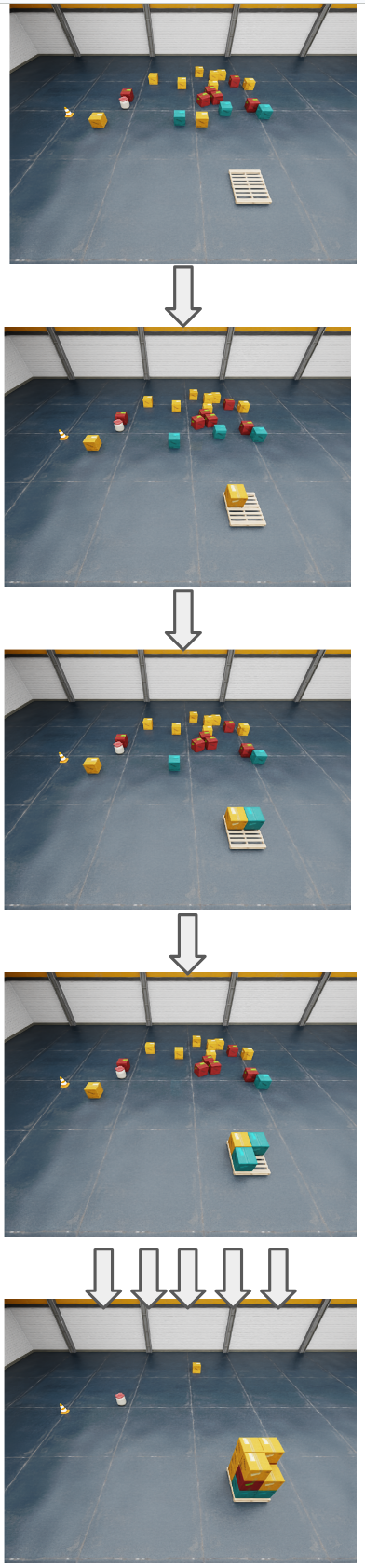}%
        \\[2pt]
        \scriptsize Goal: ``Stack the boxes with maximum height of 3, only stacking when necessary."
    \end{minipage}\hfill
    \begin{minipage}{0.25\textwidth}
        \centering
        \includegraphics[width=\linewidth,trim=0 0 0 1cm,clip]{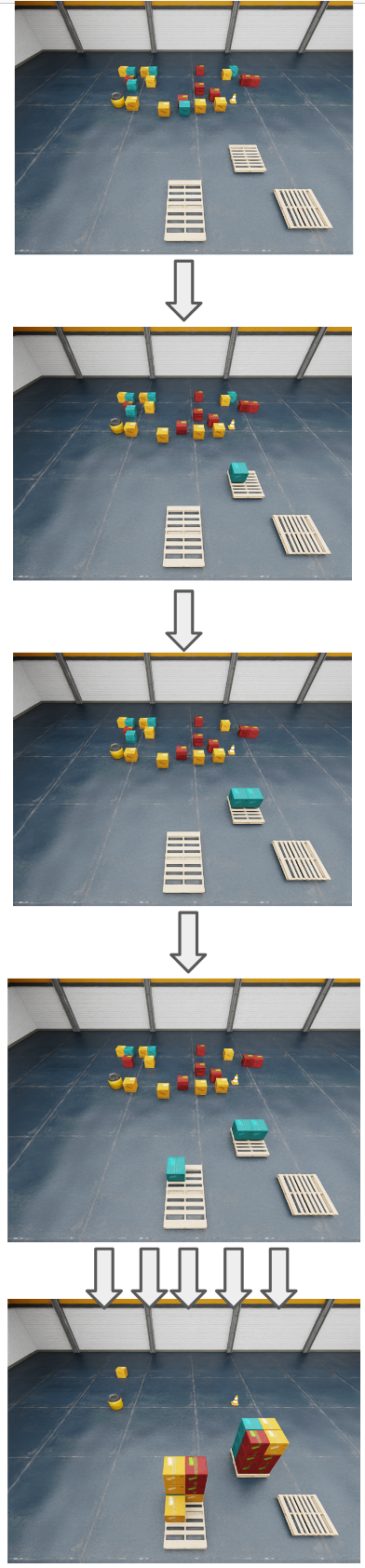}%
        \\[2pt]
        \scriptsize Goal: ``Create stacks of boxes up to height 3, starting with blue boxes."
    \end{minipage}\hfill
    \caption{Stepwise evolution of the scenes upto the penultimate step.}
\end{figure}

\pagebreak

\subsection{Baseline prompt}

\begin{quote}
\textit{You are a precise vision-language planner for warehouse box rearrangement.}

\textit{For each scene, you receive 5 images:}
\begin{itemize}
    \item \textit{Image 1: side view A of the scene}
    \item \textit{Image 2: side view B of the scene}
    \item \textit{Image 3: side view C of the scene}
    \item \textit{Image 4: side view D of the scene}
    \item \textit{Image 5: top-down (bird's-eye) view of the entire scene}
\end{itemize}

\textit{You also receive a natural-language instruction describing the goal of the rearrangement task.}

\textit{Your job is to:}
\begin{enumerate}
    \item \textit{Use all 5 images to understand the 3D layout and the goal.}
    \item \textit{Decide exactly one pickup object (a box) and exactly one putdown location (a pallet/shelf cell, stack top, or floor cell) that should be executed next to make progress toward the goal.}
    \item \textit{Describe both the pickup and putdown purely in terms of what is visible in the top-down image (Image 5) so that a separate pointing model can localize them on that image.}
\end{enumerate}

\textit{The top-down image (Image 5) is always the reference frame for your descriptions. Do NOT describe 3D coordinates or robot motions. Do NOT refer to side views in your final descriptions; side views are only for your own internal reasoning.}

\medskip
\textbf{\textit{Location Description Style (No Absolute Directions)}}

\textit{You MUST NOT use the words ``left'', ``right'', ``front'', ``back'', or any variants of them in your output.}

\textit{Instead, describe locations using:}
\begin{itemize}
    \item \textit{Proximity relations: ``closest to X'', ``furthest from X'', ``near X'', ``adjacent to X''}
    \item \textit{Between-ness: ``between X and Y'', ``in the gap between X and Y''}
    \item \textit{Center/middle: ``near the center of the pallet'', ``closest to the center of the pallet''}
    \item \textit{Pattern/structure references: ``the only pallet that has a red stack'', ``the pallet that contains exactly one red box''}
\end{itemize}

\textit{Boxes can only be placed directly on the pallets, or on top of boxes already on the pallets. Boxes cannot be placed on the floor.}

\textit{The boxes on the pallets can be arranged in a grid of $2{\times}2$ or $2{\times}3$ cells. Refer to putdown locations in relation to other boxes or pallet structure (e.g., ``corner of the pallet closest to the blue stack'').}

\textit{For every description, include:}
\begin{enumerate}
    \item \textit{Object type (box, pallet).}
    \item \textit{Color of the box, if applicable.}
    \item \textit{Supporting surface (which pallet, which shelf, or floor region), described by its visible contents or structure.}
    \item \textit{Local spatial relation expressed via ``closest to X'', ``between X and Y'', ``near X'', ``adjacent to X'', or ``near the center'', but never using any absolute directions.}
\end{enumerate}

\textit{Be unambiguous: your description should clearly correspond to a single location in the top-down image. If multiple items match, make the description more specific using additional relations. You must always choose exactly one pickup and exactly one putdown. If you truly cannot decide a valid next action, set both descriptions to ``UNDECIDABLE''.}

\medskip
\textbf{\textit{Output Format}}

\textit{Output ONLY a JSON object in this exact format (no extra text, no explanations, no markdown):}

\begin{verbatim}
{
  "pickup_description": "<description of pickup box in top-down image>",
  "putdown_description": "<description of putdown location in top-down image>"
}
\end{verbatim}

\textit{Each description should be one sentence, focused on spatial grounding in the top-down view. Describe putdown locations precisely in terms of proximity to edges, vertices, stacks, or other visible scene markers---never use the word ``cell''. Whenever moving a box, select the putdown location that is viable but also minimizes the distance from the pickup location.}
\end{quote}

\end{document}